# A Gait Foundation Model Predicts Multi-System Health Phenotypes from 3D Skeletal Motion

Adam Gabet[1,2], Sarah Kohn[1,2], Guy Lutsker[1,2], Shira Gelman[1,2], Anastasia Godneva[1,2], Gil Sasson[1,2], Arad Zulti[1,2], David Krongauz[1,2], Rotem Shaulitch[1,2], Assaf Rotem[4,5], Ohad Doron[4], Yuval Brodsky[4], Adina Weinberger[1,2], Eran Segal[1,3]*

**Author affiliations**

[1] Department of Computer Science and Applied Mathematics, Weizmann Institute of Science, Rehovot, Israel.
[2] Department of Molecular Cell Biology, Weizmann Institute of Science, Rehovot, Israel.
[3] Mohamed bin Zayed University of Artificial Intelligence, Abu Dhabi, UAE.
[4] Newton VR LTD., Tel Aviv-Yafo, Israel
[5] BioPilot AI, Newton, MA, USA

**Abstract**

Gait is increasingly recognized as a vital sign, yet current approaches treat it as a symptom of specific pathologies rather than a systemic biomarker. We developed a gait foundation model for 3D skeletal motion from 3,414 deeply phenotyped adults, recorded via a depth camera during five motor tasks. Learned embeddings outperformed engineered features, predicting age (Pearson r = 0.69), BMI (r = 0.90), and visceral adipose tissue area (r = 0.82). Embeddings significantly predicted 1,980 of 3,210 phenotypic targets; after adjustment for age, BMI, VAT, and height, gait provided independent gains in all 18 body systems in males and 17 of 18 in females, and improved prediction of clinical diagnoses and medication use. Anatomical ablation revealed that legs dominated metabolic and frailty predictions while torso encoded sleep and lifestyle phenotypes. These findings establish gait as an independent multi-system biosignal, motivating translation to consumer-grade video and its integration as a scalable, passive vital sign.

*Corresponding author:

Prof. Eran Segal, Department of Computer Science and Applied Mathematics, Weizmann Institute of Science, Rehovot, Israel, Tel: 972-8-934-3540, Fax: 972-8-934-4122, Email: eran.segal@weizmann.ac.il, ORCID: 0000-0002-6859-1164

## Introduction

Gait speed is increasingly recognized as a "vital sign" of overall health status [1]. Pooled analyses across nine cohorts (n = 34,485) have shown that gait speed independently predicts survival in older adults with accuracy comparable to complex multi-variable clinical models [2]. Beyond survival, specific gait impairments have been associated with conditions ranging from depression [3] to diabetic neuropathy [4] and neurodegenerative disease. [5]

Yet despite this clinical significance, quantitative gait analysis has remained largely focused on detecting specific pathologies such as identifying the tremor of Parkinson's disease [6], the asymmetry of post-stroke hemiparesis [7], or the patterns associated with fall risk [8]. Consequently, movement is typically treated as a symptom of dysfunction, not as a standalone biosignal for systemic health.

A key limitation impeding broader clinical utility is methodological. Most gait studies rely on engineered summary statistics, such as average cadence, stride length, and step-time variability, which reduce complex spatiotemporal sequences to a handful of scalars [9]. While clinically validated and interpretable, this reductionist approach may discard fine-grained temporal dynamics and inter-joint coordination that encode additional systemic physiological states. [10]

A systematic review of over 2,700 instrumented gait analysis studies found that more than 99% addressed only technical or condition-specific diagnostic efficacy [11]. To date, no prior work has systematically evaluated whether the full complexity of human gait can predict a broad panel of health phenotypes spanning multiple body systems in a large, deeply phenotyped healthy cohort.

Self-supervised learning of gait representations has recently shown promise for biomechanical estimation [12], clinical severity scoring [13], and diagnostic biomarker discovery [14]. However, these efforts have primarily focused on single pathologies or specific biomechanical tasks, leaving the systemic possibility of a phenome-wide gait foundation model unexplored.

Here, we adopted the Masked Autoencoder (MAE) paradigm [15], extended to skeleton sequences [16], to learn gait embeddings from 3D skeletal motion captured during five standardized motor tasks across 351 hours 3,414 adults from the Human Phenotype Project (HPP) [17]. To our knowledge, this represents the largest 3D skeletal gait dataset by subject count. Critically, movement activity was recorded with a single depth camera within a commercially available clinical movement laboratory, suggesting that meaningful health signals could be extracted from accessible recording setups.

We evaluated these embeddings against a comprehensive set of phenotypic targets spanning 18 body systems, including cardiovascular function, serum metabolomics, sleep, bone density, liver health, and mental health indices. By benchmarking against demographic and anthropometric baselines, specifically: age, body mass index (BMI), and visceral adipose tissue (VAT) [18], we demonstrate that gait embeddings provide independent predictive information across metabolic, cardiovascular, hematopoietic, and psychiatric phenotypes. Furthermore, anatomical ablation analysis reveals biologically interpretable patterns linking specific body regions to distinct health outcomes, establishing gait as a multi-system biomarker of human health.

## Results

### Deep phenotyping of the study cohort

We analyzed data from the Human Phenotype Project (HPP), a longitudinal study linking deep physiological profiling with multi-omics data. The study population consisted of 3,414 adults (aged 20–79 years) of predominantly Ashkenazi Jewish descent, residing in Israel. Participants were generally healthy at enrollment, with exclusion criteria applied for severe active disease (**Methods**).

This cohort includes extensive clinical, behavioral, physiological and multi-omics profiling categorized here into 18 body systems (Figure 1a). We focused on analyzing multi-activity gait recording data from 3,414 participants to investigate their associations with concurrent phenotypic measurements. Throughout this work, we use 'gait' broadly to encompass the full movement assessment protocol, which includes both locomotor tasks and complementary postural and functional motor assessments. The final analytical cohort included 1,652 males and 1,762 females with a mean age of 52.5 ± 10.1 years and mean BMI of 25.7 ± 3.8 kg m$^{-2}$ (**Extended Table 1**).

### Reference values and limitations of traditional gait descriptors

Gait activities consisted of five standardized motor tasks, recorded using the Movement Lab (Newton Tech), a depth camera-based movement analysis system, designed to probe complementary aspects of locomotor and postural control (**see Methods, Data Acquisition**).

### Data and model exploration

Analysis of these conventional gait descriptors revealed pronounced sexual dimorphism. During fixed-speed treadmill walking, females exhibited higher cadence (100.8 vs. 94.4 steps min$^{-1}$) and shorter stride length (0.99 vs. 1.06 m) compared with males. Functional task performance also differed by sex, with females demonstrating shorter chair-rise times during sit-to-stand (664 vs. 716 ms), whereas males exhibited larger anteroposterior sway amplitudes during the Romberg test (307 vs. 282 mm).

Despite their ability to capture gross kinematic differences, these engineered summary metrics showed limited utility for biological inference. Models trained exclusively on traditional gait descriptors failed to explain substantial variance in basic demographics such as chronological age and body mass index (BMI) (**Figure 2**).

Extensive benchmarking across the full phenotypic panel revealed that these descriptors yielded modest predictive gains above a basic demographic baseline (age, BMI) in a limited number of body systems (**Extended Figure 2**). These results indicate that while conventional gait descriptors capture some health-relevant variation, their sensitivity remains limited compared with the learned representations described below.

**Self-supervised learning of gait dynamics**

To capture the rich temporal and spatial structure of human movement, we employed a representation learning approach based on a dual-stream spatiotemporal transformer architecture trained directly on raw 3D skeletal motion sequences. Model training was performed using a denoising masked autoencoding objective, in which the network reconstructed partially observed motion sequences under combined spatial (joint-level) and temporal masking (**Methods - Deep learning section**). The trained model achieved a low reconstruction error (mean per joint position error of 0.008), corresponding to an average joint position error of approximately 8 mm, indicating accurate recovery of masked gait dynamics. (**Figure 1c**)

From the trained network, we extracted compact gait embeddings summarizing both spatial configuration and temporal coordination patterns across joints. These embeddings enabled robust prediction of chronological age across individual motor tasks. Among single-task models, treadmill walking at 3 km h$^{-1}$ achieved the strongest performance (Pearson r = 0.61±0.003), while other tasks such as the sit-to-stand task remained predictive (Pearson r = 0.53±0.002; **Figure 2c,d**). To integrate complementary information across tasks, we constructed a Gait Fusion model by averaging predictions derived from each activity-specific embedding. This ensemble substantially improved age prediction performance (male: Pearson r = 0.691±0.002, female: Pearson r = 0.682±0.002), significantly outperforming models based on conventional gait descriptors combined with height (male: Pearson r = 0.455±0.007, female: Pearson r =0.512±0.006).

The same representations captured body composition with high fidelity. Regression on BMI using the fused model yielded strong correlations in both sexes (male: Pearson r = 0.882±0.0, female: Pearson r = 0.898±0.002, Figure 2e,f). To assess relevance to metabolically meaningful adiposity, we further regressed gait embeddings against visceral adipose tissue (VAT) area, achieving male Pearson r = 0.817±0.001, female Pearson r =0.795±0.002, compared with Pearson r =0.518±0.010 and 0.467±0.007 respectively using conventional gait features and height (Figure 2g,h). Gait embeddings also discriminated sex with near-perfect accuracy (ensemble AUC = 0.99±0.002; >0.95 for all individual tasks), motivating the sex-stratified analyses reported below. Additionally, Gait embeddings captured individual-specific signatures that were temporally stable across repeat visits (**Supplementary Figure 2**).

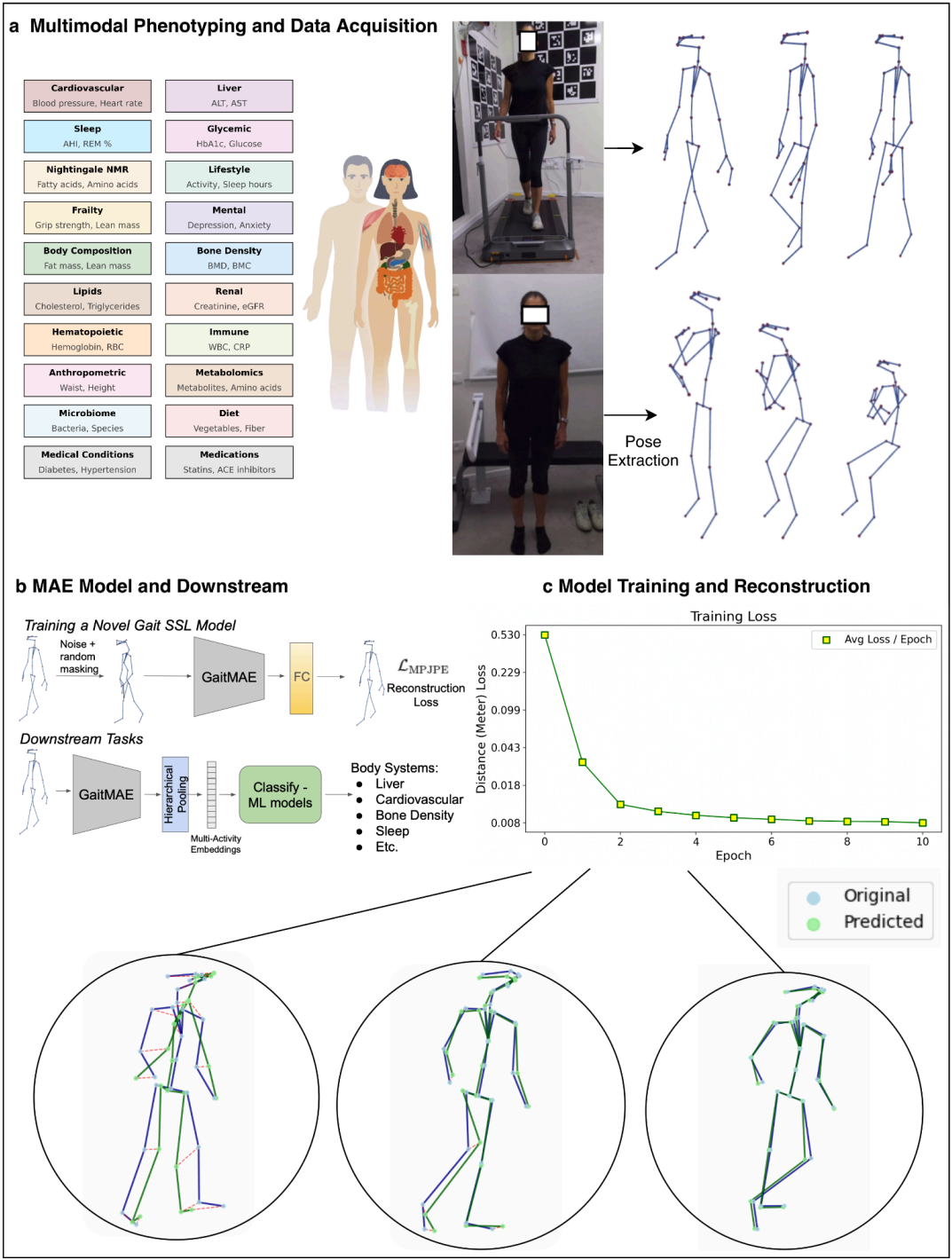

**Figure 1**: **Overview of the Gait MAE Model, data acquisition, and training pipeline. a,** Multimodal phenotyping and gait acquisition. Left: The study cohort (Human Phenotype Project) underwent deep profiling across 18 distinct physiological systems (color-coded categories), ranging from cardiovascular and metabolic markers to mental health and lifestyle indices. Right: Gait data were acquired using a single Azure Kinect depth camera during five standardized motor tasks (for example, treadmill walking and sit-to-stand). Raw depth data was processed to extract 3D skeletal pose sequences. **b,** Self-supervised architecture and downstream evaluation. Top: The GaitMAE model is pre-trained using a denoising masked autoencoder objective. Input skeleton sequences are corrupted with Gaussian noise and random spatio-temporal masking; the model is trained to reconstruct the original motion,

minimizing the mean per-joint position error down to 0.008 (kin to 8mm error). Bottom: For downstream tasks, representations from the frozen encoder undergo hierarchical pooling to generate subject-level multi-activity embeddings, which are then used to train standard machine learning models for phenotype prediction. **c,** Model training and reconstruction performance. Top: Training loss curve (mean distance error in meters) showing convergence to a reconstruction error of approximately 8 mm. Bottom: Qualitative visualization of reconstruction fidelity, displaying the overlap between ground truth (Original; blue) and model-reconstructed (Predicted; green) skeletons. ACE, angiotensin-converting enzyme; ALT, alanine transaminase; AST, aspartate transaminase; BMI, body mass index; BMD, bone mineral density; CRP, C-reactive protein; eGFR, estimated glomerular filtration rate; HbA1c, glycated hemoglobin; HDL-C, high-density lipoprotein cholesterol; LDL-C, low-density lipoprotein cholesterol; RBC, red blood cells; REM, rapid eye movement; WBC, white blood cells.

**Gait embeddings alone predict phenotypes across 18 body systems**

To identify key associations within other body systems, we explored the prediction power of our Gait Fusion model on features from all other body systems. This analysis simulates a passive screening setting in which only a gait recording is available, with no accompanying clinical or demographic information.
We extracted gait feature embeddings for each subject in our study cohort and trained with linear and non-linear models for each phenotype. In total, we found 1,980 significant associations out of 3,210 phenotypes tested after False Discovery Rate (FDR) correction.

Predictive performance was strongest for structural and frailty metrics: body composition, bone density, and frailty metrics (Lean mass and hand grip strength) came out as most significant. In addition, engineered gait features were also strongly predicted, serving as a technical validation. Sleep and Cardiovascular measures were predicted to a substantial extent compared to other body systems.

Notably, organ-specific function was also captured, with liver function features such as liver sound speed and liver elasticity were relatively strongly predicted. Our Lifestyle features come from questionnaires and as such remain elusive to high predictions but gait most strongly predicted walking habits and exercise out of all lifestyle questions (**Figure 3**). In addition, Gait exhibited strong connections to the molecular level demonstrating significant associations with cholesterol, hemoglobin and various blood tests connections.

Stratification by sex confirmed that these predictive trends were broadly concordant across sexes, though notable divergences emerged (**Extended Figure 1**). Frailty, body composition, and dietary features were more strongly predicted in males, whereas cardiovascular and liver phenotypes showed higher predictive performance in females.

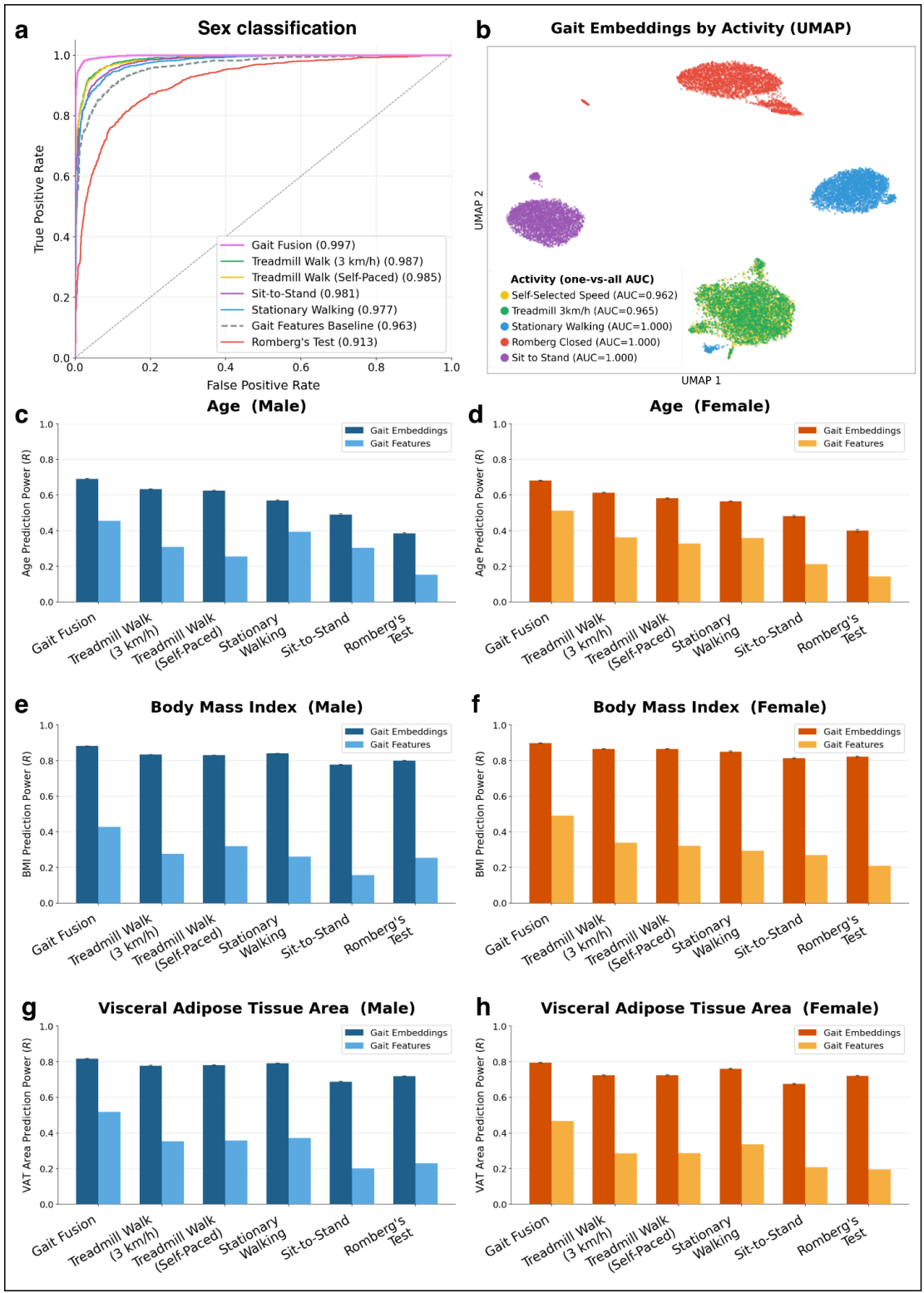

**Figure 2 | Gait embeddings encode biological characteristics with high fidelity.**

a, Sex classification performance evaluated by ROC curves across individual motor tasks and the Gait Fusion ensemble. All embedding-based models exceed AUC = 0.91, with Gait Fusion achieving AUC = 0.997. The engineered gait features baseline (dashed) achieves AUC = 0.963. b, UMAP projection of gait embeddings colored by motor task, showing distinct activity-specific clusters. One-vs-all AUC values confirm task separability (AUC ≥ 0.962). c,d, Chronological age prediction (Pearson r) for male (c) and female (d) participants, comparing learned gait embeddings against engineered gait features (cadence, stride length, sway metrics, and height) across individual tasks and the Gait Fusion ensemble. e,f, BMI prediction for male (e) and female (f) participants. g,h, Visceral adipose tissue (VAT) area prediction for male (g) and female (h) participants. Across all targets and both sexes, learned embeddings (dark bars) substantially outperform engineered features (light bars), with the Gait Fusion ensemble consistently achieving the highest correlation. All predictions were evaluated using 5-fold nested cross-validation repeated across 15 random seeds (n = 75 train–test splits), with sex stratification applied throughout (Methods).

**Adjusting for Covariates Across Body System-Level Associations**

While direct prediction of phenotypes supports the utility of gait for accessible screening, these results raise a critical question: do gait embeddings capture distinct biological signals, or do they merely proxy well-established confounders? Our model predicts age, BMI, and sex with high accuracy (**Figure 2a–b**), and these variables are known to drive a wide range of health phenotypes [19,20]. Moreover, VAT, a clinically meaningful marker of metabolic risk[21], is also predicted with high fidelity. It is therefore plausible that the observed phenotypic associations are mediated entirely through these factors.

To address this, we employed a predictive gain approach: for each phenotype, we compared the predictive performance of demographic and anthropometric confounders alone (age, BMI, height [22], VAT) against confounders combined with gait embeddings extracted from the MAE model. Both models were trained with linear and non-linear models and were evaluated within a nested cross-validation framework . All analyses were further stratified by sex to prevent sex-linked gait variation from being conflated with phenotype-relevant signals. Performance was evaluated using Pearson r, and the improvement (Δr) attributable to gait is reported across body systems (**Methods - predictive models**).

Gait embeddings consistently yielded incremental predictive power across body systems, indicating that movement encodes health information orthogonal to body habitus and age (Figure 4). In males, gait embeddings significantly improved prediction in all 18 body systems; in females, 17 of 18 systems showed significant gains, with renal function as the sole exception. The largest median improvements were observed in hematopoietic markers, mental health phenotypes, and frailty indices. Metabolomics and liver markers also showed substantial gains, while bone density, despite a moderate median improvement, exhibited the widest spread, with several individual phenotypes reaching Δr values above 0.15. Cardiovascular and body composition phenotypes showed more modest but consistent gains. Even systems with smaller median deltas, such as sleep and metabolomics, contained individual phenotypes with notable improvements, suggesting gait captures information relevant to diverse physiological processes. These results position gait not as a proxy for age or body habitus, but as an independent biosignal with broad clinical relevance (**Figure 4**).

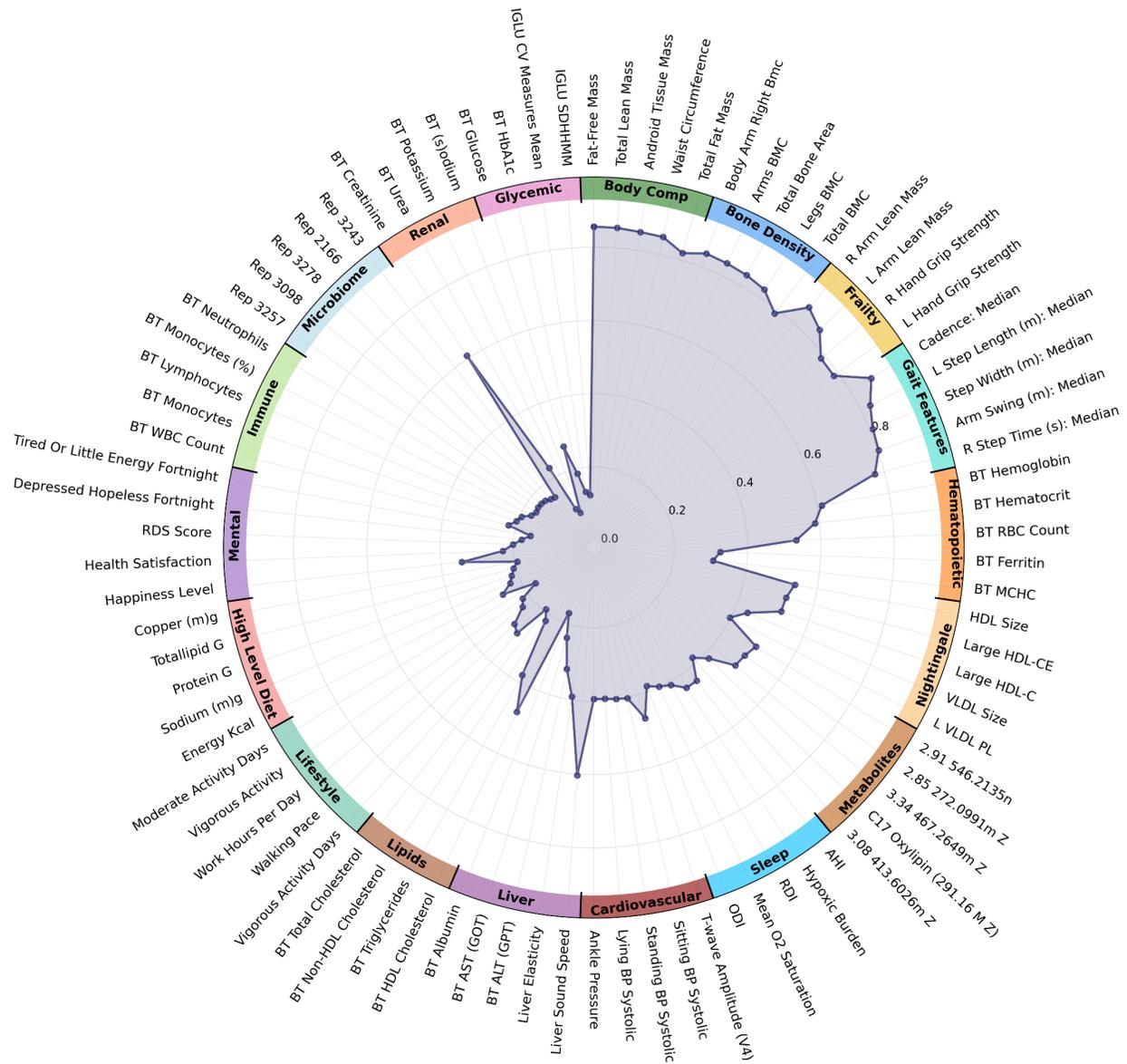

**Figure 3 | Gait embeddings independently predict phenotypes across diverse body systems.** Radar plot showing the top 5 features most strongly predicted by the Gait Fusion model for each body system (FDR-corrected $P < 0.05$). This analysis simulates a passive screening scenario in which only a gait recording is available, with no demographic, anthropometric, or clinical covariates included as model inputs, equivalent to predicting a person's health profile solely from filming them as they walk. Each slice represents a single phenotype, grouped by body system (labeled in the inner ring). The radial axis spans Pearson $r$ correlation of predictions from 0 to 1. Features within each body system are ordered clockwise. Predictions were generated using linear ridge regression on gait embeddings from the frozen GaitMAE encoder, evaluated via 5-fold nested cross-validation repeated across 15 random seeds ($n = 75$ train–test splits). AHI, apnea–hypopnea index; ALT, alanine transaminase; AST, aspartate transaminase; BMC,

bone mineral content; BT, blood test; HDL, high-density lipoprotein; HDL-C, high-density lipoprotein cholesterol; HDL-CE, high-density lipoprotein cholesterol ester; IGLU, indices of glycemic variability; MCHC, mean corpuscular hemoglobin concentration; ODI, oxygen desaturation index; RDI, respiratory disturbance index; RBC, red blood cells; RDS, recent depressive symptoms; VLDL, very-low-density lipoprotein; WBC, white blood cells.

**Gait signatures correlate with clinical diagnoses and medication use**

To evaluate the clinical relevance of gait beyond continuous metabolic markers, we examined the associations between gait embeddings and a broad panel of medical diagnoses and medication usage. Analyses were stratified by sex and performed using logistic regression models rigorously adjusted for age, BMI, and VAT (**Methods**).

Gait embeddings contributed significant independent predictive information across a broad spectrum of conditions, substantially exceeding the baseline in nearly all cases (**Figure 5**). In male participants, the strongest predictive gains appeared for depression, glaucoma, insomnia, back pain, and peptic ulcer disease. These associations were corroborated by predictable usage of related medication categories, including proton pump inhibitors (PPIs), antihistamines, and antidepressants. In female participants, the ensemble model showed particularly strong performance for irritable bowel syndrome (IBS), peptic ulcer disease, psoriasis, headache, depression, and diabetes. Predictable medication usage again mirrored these conditions, with strong associations found for adrenergic inhalants, iron supplements, and hormone antagonists.

Sex-specific patterns were evident across both conditions and medications. Male participants showed unique associations with heart valve disease, vitiligo, and sinusitis, whereas female participants exhibited distinct associations with gynecological conditions, retinal detachment, and insomnia. Medication patterns mirrored this divergence, antidepressant use was predictable from gait in both sexes, while iron supplement use was specific to female participants [23]

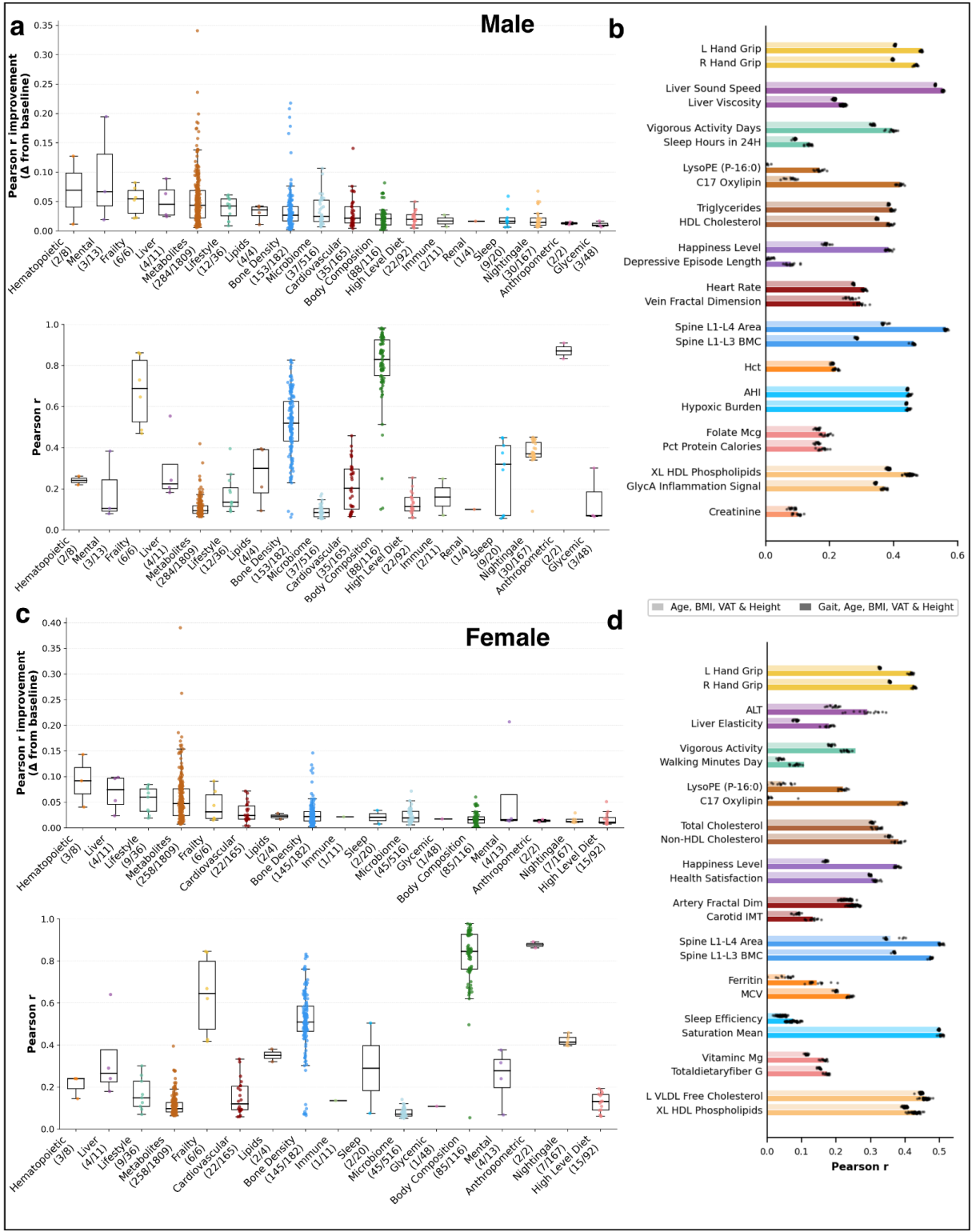

**Figure 4 | Gait embeddings provide independent predictive information beyond demographic and anthropometric confounders across body systems.**

**a,c,** Box and swarm plots (center, median; box, interquartile range (IQR); whiskers, 1.5× IQR) showing the predictive performance of gait embeddings beyond a demographic baseline for male (a) and female (c) participants. Bottom panels show the median Pearson correlation r between predicted and actual values for models combining gait embeddings with age, BMI, VAT and height. Top panels show the predictive gain (Δr) attributable to gait embeddings relative to the baseline model trained on age, BMI, VAT and height alone. Predictive power was evaluated using 5-fold nested cross-validation repeated across 15 random seeds (n = 75 train–test splits). The number of features with significantly improved predictions when adding gait embeddings (two-sided t-test, FDR-corrected P < 0.1), corresponding to the number of dots (n) included in each distribution, is shown for each body system in brackets on the x axis as (n/total number of features in the body system). Body systems are ordered from left to right by descending median Δr (top panels). **b,d,** Horizontal bar and dot plots (n = 15 seeds; bar, mean; bar, mean; whiskers, s.d) comparing the performance of the baseline model (age, BMI, VAT and height; light bars) versus the full model (gait embeddings combined with age, BMI, VAT and height; dark bars) for selected features per body system, for male (**b**) and female (**d**) participants. Features were selected to highlight clinically and physiologically notable associations across body systems, prioritizing phenotypes with substantial predictive gain (Δr) and moderate baseline performance (Pearson r < 0.6). Features are grouped and color-coded by the body system. Individual cross-validation Pearson r values are plotted as dots. All analyses were sex-stratified throughout. AHI, apnea–hypopnea index; ALT, alanine transaminase; BMC, bone mineral content; ECG, electrocardiogram; HCT, hematocrit; HDL, high-density lipoprotein; MCHC, mean corpuscular hemoglobin concentration; MCV, mean corpuscular volume; RDS, recent depressive symptoms; VLDL, very-low-density lipoprotein; WBC, white blood cells; S1P, sphingosine-1-phosphate.

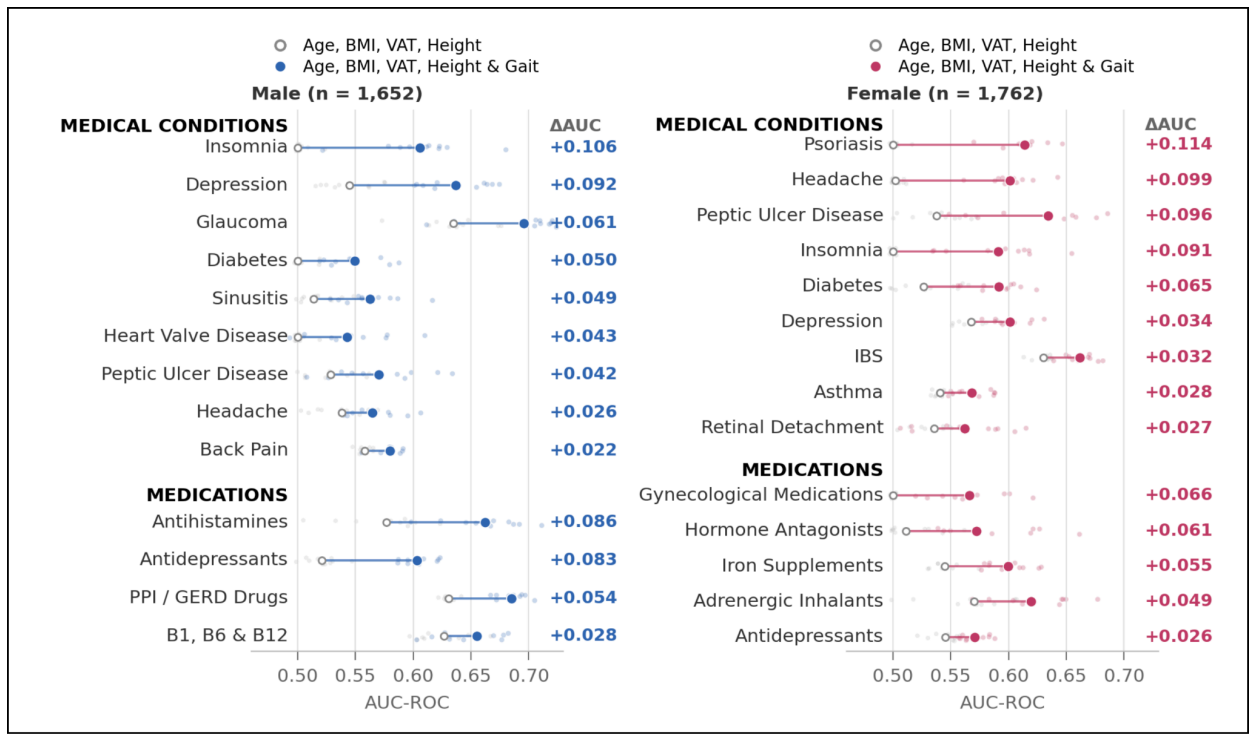

**Figure 5 | Gait embeddings improve prediction of clinical diagnoses and medication use beyond demographic and anthropometric baselines.**

**a,b** Dumbbell plots comparing AUC-ROC for models trained on covariates alone (age, BMI, VAT, height, open circles) versus models combining covariates with gait embeddings (filled circles) for male (**a**) and female (**b**) participants. Top sections show medical conditions; bottom sections show medication categories. The predictive gain (ΔAUC) attributable to gait is annotated for each target. Conditions and medications were selected as the top entries by ΔAUC within each sex. All models used logistic regression with ridge regularization, evaluated via 5-fold nested cross-validation repeated across 15 random seeds ($n = 75$ train–test splits). Significance was assessed by two-sided *t*-test comparing full versus baseline model AUC distributions (FDR-corrected $P < 0.1$). Analyses were sex-stratified throughout. IBS, irritable bowel syndrome; PPI, proton pump inhibitor; GERD, gastroesophageal reflux disease.

**Interpretability Analysis**

To investigate which anatomical regions drive the model's health predictions, we conducted body-part ablation studies. We defined four joint groups - head, arms, torso, and legs; and computed an importance score for each group per health outcome by combining two complementary ablation strategies: measuring the performance drop when masking a group from the full skeleton , and measuring each group's isolated predictive contribution relative to other groups. This analysis was restricted to treadmill walking activities, as comparing anatomical importance across task types would conflate task-specific biomechanical demands with health-relevant signals. Scores were normalized to [0, 1] and averaged across the top features per body system (**see Methods**). Full results are presented in Figure 6; here we highlight the most notable patterns.

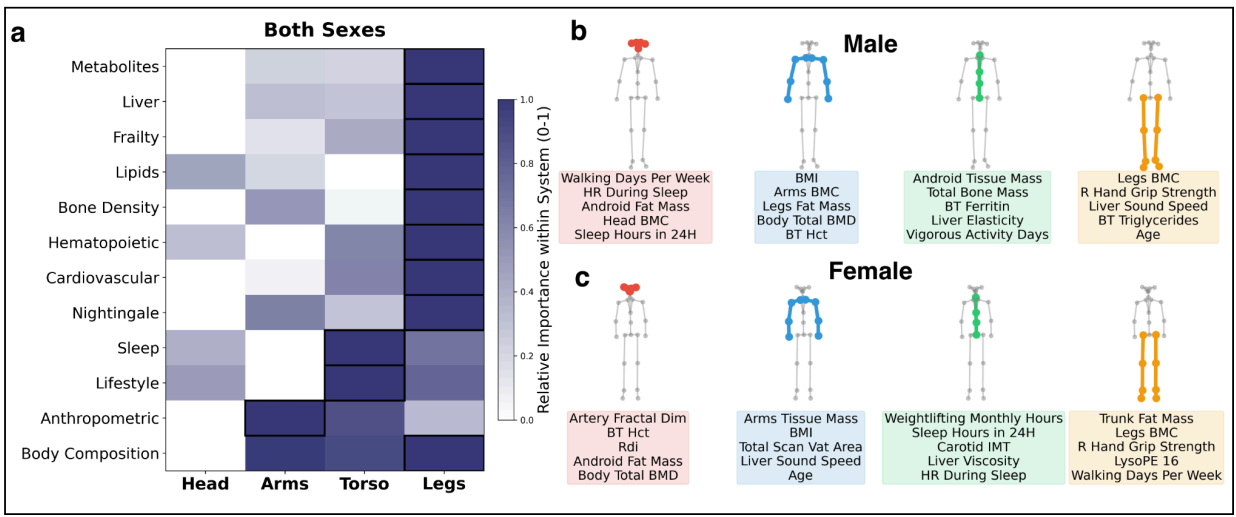

**Figure 6 | Interpretability of gait embedding predictions reveals anatomically specific health associations.**

**a,** Heatmap showing the mean importance score of the top 10 significantly predicted labels per body system across four anatomical joint groups (head, arms, torso, legs). Importance scores were computed by combining two complementary ablation strategies: the performance drop when masking a group from the full skeleton and each group's isolated predictive contribution relative to other groups. Scores were normalized to [0, 1] and averaged across the top 10 features per body system. Sex was included as a covariate to capture importance patterns across both sexes. Legs emerged as the dominant predictive region across most body systems, while sleep and lifestyle phenotypes were most strongly associated with torso dynamics. **b,c,** Sex-stratified anatomical attribution for five representative health outcomes per joint group in male (**b**) and female (**c**) participants. Each skeleton highlights the joint group (color-coded) most important for the listed phenotypes. BMC, bone mineral content; BMD, bone mineral density; BT, blood test; ECG, electrocardiogram; Hct, hematocrit; MCHC, mean corpuscular hemoglobin concentration; R, right; VAT, visceral adipose tissue area.

The legs emerged as the most important joint group for 9 of 12 body systems (**Figure 6a**) and were the top-ranked region for more individual phenotypes than any other group in both sexes (64 of 159 labels in males; 70 of 159 in females). This is consistent with the central role of lower-limb kinematics in conventional gait analysis [24] [11]. Beyond standard locomotor metrics, leg dynamics were important for predicting general frailty markers such as grip strength, hematopoietic parameters including hemoglobin and hematocrit, and serum lipids such as triglycerides, in both sexes. Leg importance also extended to behavioral phenotypes, with self-reported weekly walking duration among the lifestyle features most strongly attributed to this region.

As expected, anthropometric and body composition targets showed distributed importance across all four anatomical regions (**Figure 6a**), reflecting the whole-body nature of these measurements.

An unexpected finding was that the torso joint group carried the highest importance for both sleep and lifestyle body systems (**Figure 6a**). Among the torso-associated phenotypes were ECG T-wave amplitude, sleep hours, heart rate during sleep, and weightlifting duration (**Figure 6b,c**). The association with weightlifting time is notable given that this variable was derived from wrist-worn accelerometry, yet its prediction was driven by trunk movement patterns rather than arm kinematics, suggesting that resistance training leaves a lasting imprint on trunk postural dynamics during gait. The torso may therefore serve as a convergence point in which both sleep quality and exercise habits manifest as measurable changes in trunk dynamics during gait.

Several phenotypes showed distinct anatomical attributions between sexes (**Figure 6b,c**). Most notably, chronological age prediction was driven predominantly by leg dynamics in males but by arm and torso dynamics in females. Bone mineral density (BMD) also exhibited sex-specific patterns: in females, both total and regional BMD were most strongly associated with the head joint group, whereas in males BMD importance was distributed more evenly across regions. These sex-divergent attribution patterns suggest that the biomechanical signatures of aging and skeletal health are not uniform across sexes, an observation we explore further in the Discussion.

Taken together, the anatomical attribution analysis reveals that gait embeddings do not rely on a single movement signature but instead encode region-specific biomechanical patterns that map onto distinct physiological systems. The observation that different body regions predict different health domains suggests that targeted motor assessments, such as trunk-focused tasks for sleep and lifestyle screening or lower-limb protocols for metabolic and frailty markers, could be designed to improve the sensitivity of gait-based health monitoring for specific clinical applications.

## Discussion

For this study, we analyzed 17,589 gait sequences from 3,414 deeply phenotyped adults in the Human Phenotype Project, applying self-supervised deep learning to extract representations from 3D skeletal motion captured with a single depth camera [14]. Gait biomarker predictions span 18 body systems and survive adjustment for age, BMI, visceral adipose tissue, and height. As such, we propose a paradigm shift: from analyzing gait as a symptom revealer through deviations in engineered summary statistics [25], to treating gait as an independent multi-system biomarker revealed through deep neural networks.

This shift is grounded in a fundamental representational difference. While conventional gait descriptors have proven clinically valuable [26], and while engineered features in our own cohort successfully predicted several biological measures, these descriptors reduce complex spatiotemporal sequences to a handful of scalars, potentially discarding inter-joint coordination and temporal dynamics that encode systemic physiological state. In our cohort, models trained on engineered features yielded significant predictive gains in a some of body systems but across the majority of the body systems failed to improve prediction beyond a basic demographic baseline (Extended Figure 2), whereas learned embeddings yielded significant gains in all 18 systems in males and 17 of 18 in females (Figure 4). The performance gap was most striking for targets with no obvious kinematic correlate, including liver, hematopoietic markers, lifestyle and mental health phenotypes, suggesting that the information captured by deep representations is qualitatively different from what summary statistics can access.

A key advantage of our approach is its clinical accessibility. All gait recordings were acquired using a single front-facing depth camera via a commercially available movement analysis system (Movement Lab, Newton Tech), requiring no wearable sensors, force plates, or specialized gait laboratory infrastructure. This stands in contrast to most instrumented gait analysis studies, which rely on multi-camera motion capture systems or pressure-sensitive walkways that limit deployment to specialized centers [11]. Moreover, architectures closely related to ours have demonstrated successful reconstruction of 3D pose from monocular 2D video [27], raising the possibility that comparable health-relevant embeddings could be extracted from standard smartphone or clinic-mounted cameras [28,29]. If validated, this would enable gait-based health screening at population scale with effectively zero marginal hardware cost.

A central finding of this work is that gait embeddings provide predictive information independent of age, body mass index (BMI), visceral adipose tissue (VAT), and height. Our model predicts VAT with high fidelity ($r = 0.82$), yet gait continues to improve phenotype prediction even after VAT is included as a covariate, suggesting that gait captures aspects of metabolic state not reducible to body composition alone. Given that BMI fails to distinguish metabolically healthy from metabolically unhealthy individuals at similar body weight [30], gait may provide a more functionally integrated readout of systemic metabolic status.

Among the most notable examples is the prediction of liver elasticity and sound speed beyond the VAT-adjusted baseline. That gait adds predictive value for these markers is striking, as VAT is itself one of the strongest non-invasive correlates of Metabolic Dysfunction-Associated Steatotic Liver Disease (MASLD) severity [31]. A plausible explanation lies in the bidirectional relationship between liver disease and skeletal muscle: sarcopenia and MASLD share pathophysiological pathways including insulin resistance, chronic inflammation, and altered myokine signaling, and their co-occurrence confers increased mortality beyond either condition alone [32]. Consistent with this, advanced liver fibrosis has been independently associated with decreased gait speed [33], and a recent meta-analysis of over 288,000 participants confirmed reduced gait speed in MASLD patients [34]. Our findings extend these observations from simple gait speed in clinical populations to deep learning-derived gait features in a generally healthy cohort, suggesting that subclinical hepatic dysfunction may be detectable from movement patterns before overt disease manifests.

The prediction of depression and insomnia beyond demographic baselines in both sexes is consistent with well-established bidirectional associations between depressive symptoms and altered gait patterns [3]. That antidepressant use was also captured by gait is noteworthy: all major classes of antidepressants, including Selective Serotonin Reuptake Inhibitors (SSRIs), Serotonin-Norepinephrine Reuptake Inhibitors (SNRIs), and tricyclic antidepressants, have been associated with movement disorders such as parkinsonism, akathisia, dystonia, and tremor, likely through indirect modulation of dopaminergic function in the basal ganglia [35,36], suggesting that pharmacological treatment may alleviate mood symptoms without fully normalizing motor behavior, or that the medications themselves introduce distinct gait alterations.

At the molecular level, gait embeddings predicted metabolites with direct relevance to neurodegenerative and inflammatory pathways. LysoPE(P-16:0), a plasmalogen-type lysophospholipid enriched in brain and myelin, was among the top gait-associated metabolites in both sexes. Plasmalogen depletion is a recognized biomarker of Alzheimer's disease and Parkinson's disease [37], raising the possibility that gait captures early neuroinflammatory changes before clinical symptoms emerge. Oxygenated C17 lipid species were also strongly associated with gait embeddings, consistent with prior work showing that 17 of 42 of tested serum oxylipins correlated with gait speed in older men, with associations only partially explained by muscle mass [38].

Beyond these, the concordance between predicted medical conditions and their corresponding medication categories provides internal validation that gait embeddings capture genuine biological signals. Gastrointestinal conditions including peptic ulcer disease and IBS were predicted alongside proton pump inhibitor use in both sexes, likely reflecting altered trunk mechanics due to abdominal discomfort.In female participants, iron supplement use was predicted alongside hematopoietic markers such as ferritin and Mean Corpuscular Volume. While each marker individually reflects multiple physiological processes, their joint prediction alongside iron supplement use suggests a convergent signal related to iron deficiency anemia, a condition whose motor manifestations include fatigue, reduced exercise tolerance, and diminished oxygen-carrying capacity, and which disproportionately affects women of reproductive age, with an estimated 825 million women affected globally [39]. The prediction of asthma and adrenergic inhalant use in females, combined with strong gait associations with self-reported exercise duration, suggests that respiratory health state leaves a measurable imprint on movement patterns, potentially through reduced cardiopulmonary reserve during locomotion. Along similar lines, gait predicted diabetes as a clinical diagnosis despite minimal association with glycemic variability, suggesting it captures the motor consequences of peripheral neuropathy - impaired proprioception and altered foot mechanics rather than glucose homeostasis per se [40,41].

Our interpretability analysis revealed that both sleep and lifestyle phenotypes were most strongly driven by trunk dynamics rather than limb kinematics, which is biologically plausible: sleep deprivation has been shown to impair postural control and increase trunk sway [42,43], and habitual physical activity improves core neuromuscular control and trunk stability [44,45], suggesting the torso may serve as a convergence point where both sleep quality and exercise habits manifest as measurable changes in gait. Beyond region-specific patterns, the anatomical attribution also revealed notable sex differences: age prediction was driven by leg dynamics in males but by arm and torso kinematics in females, consistent with evidence that men experience greater rates of lower-limb strength decline with age [46] and that the cellular mechanisms of age-related muscle atrophy differ between sexes myofiber loss in males versus myofiber atrophy in females [47,48]. Together with the sex-divergent clinical associations observed in Figure 5, these patterns suggest that sex-stratified motor assessments could improve the sensitivity of gait-based health screening.

This work contributes to a growing body of research applying deep neural networks to gait analysis [49] [12,14], and strongly underscores the value of this shift by demonstrating predictive capacity across body systems rather than within a single pathology. We acknowledge several limitations. Our cohort is composed predominantly of Ashkenazi Jewish adults residing in Israel, and generalizability to other populations remains to be established. Hardware specificity poses an additional barrier: our model was trained on Azure Kinect depth camera recordings, and adapting to other capture systems or public gait datasets [50,51] would require non-trivial domain transfer. All reported associations are cross-sectional; causal inference will require longitudinal follow-up, which the HPP's 25-year design is positioned to provide. Our interpretability analysis operates at the level of four anatomical groups, and finer-grained joint-level or temporal attribution remains a direction for future work. Finally, while we demonstrate that gait embeddings predict clinical diagnoses and medications beyond demographic baselines, prospective validation in clinical screening settings is needed to establish diagnostic utility. Future directions include longitudinal prediction of disease onset, external validation on ethnically and geographically diverse cohorts, translation to consumer-grade 2D cameras, and the design of targeted motor assessments, such as trunk-focused tasks for sleep and lifestyle screening, or lower-limb protocols for metabolic and frailty markers ,informed by the anatomical attribution patterns reported here.

## Methods

**Description of cohort**

The data presented in this paper were collected between January 2023 and December 2025, from a total of 3,414 participants aged between 20 years and 79 years, who were enrolled as part of the HPP study and who underwent gait activity analysis as part of the study. This study was approved by the institutional review board of the Weizmann Institute of Science (reference no. 1719-1), and all participants were self-assigned volunteers with informed consent. The cohort in this study is one of the largest longitudinal studies established in Israel, with a population originating from several different ancestries who reside in a relatively small geographic region and, therefore, share a relatively similar environment and habits. The population is largely composed of educated European (Ashkenazi) Jews, healthy at the time of recruitment (that is, severe medical conditions and diseases were defined as exclusion criteria), who have follow-up visits every 2 years for 25 years [18]. For the full study design, see [17]. The data include various clinical, physiological, behavioral and multi-omic profiling data, which we categorized into 18 body systems (excluding gait characteristics) representing major physi-ological systems and environmental exposures (**Figure 1**). [52] The following groups represent the baseline characteristics and system-level categories used for this work. The exact number of individuals, data points and main features available in each group can be found in Extended Table 1.

Anthropometric: includes 3 body measurements collected during the clinical visit: waist circumference, hip circumference, and body weight, representing basic anthropometric indices of body size and central adiposity. BMI, height, and detailed regional body composition measurements were excluded from this category as they were accounted for in the baseline covariate group and body composition system, respectively.

Blood Tests lipids: includes the following blood tests: high-density lipoprotein cholesterol, non-high-density lipoprotein cholesterol and triglycerides.

Body composition: includes 108 measurements of fat and lean mass assessments for the legs, arms, trunk, gynoid and android regions. These measurements are derived from dual-energy X-ray absorptiometry imaging. Additionally, anthropometric parameters, such as weight, height and hip and waist circumference, were also included. BMI was not part of this category because it was already accounted for in the baseline group.

Bone density: includes 182 measurements of the mineral content in different parts of a variety of skeletal components, based on dual-energy X-ray absorptiometry imaging.

Cardiovascular system: includes measurements from various tests: blood pressure measurements, blood pressure ratios computed using the ankle–brachial index test, arterial stiffness estimated by pulse wave velocity using a Falcon device (Viasonix) and carotid intima-media thickness computed from the carotid ultrasound using a SuperSonic Aixplorer MACH 30 (Hologic), vascular parameters averaged over both eyes computed from retinal imaging using an iCare DRSplus confocal fundus imaging system (iCare) and the Python AutoMorph package61 , as well as electric activity of the heart as captured in the 12-lead resting ECG using a PC-ECG 1200 machine (NORAV).

Frailty: includes 6 features capturing the key components of physical frailty: appendicular lean mass of the four limbs (right and left leg, right and left arm) derived from dual-energy X-ray absorptiometry and bilateral hand grip strength measured by dynamometry.

High-level diet: includes 124 measurements encompassing macronutrients, micronutrients, meal patterns, and food categories. This system integrates fundamental nutritional components, including energy intake, proteins, carbohydrates, and dietary fiber - alongside 13 vitamins, 11 minerals, and specific profiles for fatty acids and sugars. Behavioral dietary aspects are captured through 29 distinct food categories (e.g., whole grains, vegetables, and processed meats) and 4 meal pattern metrics such as fasting window and meal frequency. These features are derived from mean daily consumption values calculated from longitudinal diet logging records within a window around each research stage, using a 500 kcal daily energy threshold to ensure data reliability. Finally, macronutrient caloric ratios and lipid distribution metrics were included to provide a normalized assessment of dietary composition for predictive gait modeling.

Gut microbiome: includes relative abundances of 627 families. We used the method for metagenomic reads extraction and bacterial abundances estimation previously described [53] at the family level, in combination with the previously published improved human gut microbiome reference set for mapping [54]. Bacterial families that were not identified in at least 5% of the samples were not included. Missing data were assumed to represent missing abundance or abundances that are below the detection limit. Thus, missing data were imputed by a minimum value of 0.0001. In this work, the log10-transformed values of the resulting abundances per sample were used.

Hematopoietic system: includes the following blood laboratory test measurements, which are characteristics and components of red blood cells: ferritin, hemoglobin, mean corpuscular hemoglobin concentration, mean corpuscular volume, red blood cells, hematocrit, mean corpuscular hemoglobin and red blood cell distribution width. Immune system: includes the following immune cell measurements from complete blood count: white blood cells, absolute eosinophils, percentage eosinophils, absolute monocytes, percent- age monocytes, absolute lymphocytes, percentage lymphocytes, absolute basophils, percentage basophils, absolute neutrophils and percentage neutrophils.

Glycemic status: includes 49 parameters describing blood glucose variability computed as previously described [55]. In addition, this category includes fasting glucose and HbA1C from laboratory tests.

Lifestyle: assessed by 46 questions from questionnaires, following the structure and content of the UK Biobank [56], addressing the following topics: smoking, alcohol consumption, physical activity, employment, sleep, sun exposure and electronic device use (**full list in Supplementary Table 1**).

Liver health: assessed by parameters from a liver ultrasound and 2D shear wave elastography (2D-SWE), including measurements of viscosity, elasticity, attenuation and sound speed. In addition, the following liver enzymes and related blood tests are included: alkaline phosphatase, alanine transaminase, aspartate transaminase, total protein, total bilirubin, platelets and albumin.

Nightingale NMR: includes serum metabolic biomarkers quantified by the proton nuclear magnetic resonance ($^1$H-NMR) platform of Nightingale Health. This platform provides simultaneous quantification of 228 absolute-value-based plasma metabolites and ratios, mainly expanding detailed lipidomic profiles and adding measurements of clinically validated biomarkers, including routine lipids, lipoprotein subclass profiling with lipid concentrations within 14 subclasses, fatty-acid composition, and various low-molecular-weight metabolites such as amino acids, ketone bodies, and glycolysis metabolites [57].

Serum untargeted metabolomics (n = 1X,000) was performed via UPLC-IMS-MS (Waters SELECT SERIES Cyclic IMS) in batches of 416 injections. Processed/solvent blanks were used for background subtraction, while pooled QCs (intra-batch) and "Anchor" samples (inter-batch) corrected for analytical drift. Post-alignment, 1,809 high-quality features were retained. Three gait-associated leads, including LysoPE(P-16:0/0:0), achieved MSI Level 2 status (< 0.5 ppm mass accuracy, < 1% CCS error, and MS/MS fragments). Remaining features, such as a C_17 oxylipin-like series (C17H24O4-5), were assigned MSI Level 4 formulas via < 0.8 ppm accuracy and > 90% isotopic match.

Sleep: includes 448 characteristics derived from multi-night home sleep apnea testing using the WatchPAT device (ZOLL Itamar) [18]. Features encompass respiratory indices (apnea–hypopnea index, oxygen desaturation index, respiratory disturbance index), sleep architecture parameters (total sleep time, sleep efficiency, REM and non-REM durations), body position, snoring metrics, and heart rate dynamics during sleep.

Mental health: assessed by 35 questions from a questionnaire fol- lowing the structure and content of the UK Biobank [56] equivalent questionnaire relating to the individual's mood, satisfaction and depressive symptoms. In addition, recent depressive symptoms (RDS) score was computed as the sum of the self-reported scores of the following questions; each self-reported score is a number between 1 and 4 (1 indicates not at all, 4 indicates nearly every day) and, therefore, ranges from 4 to 16: Over the past two weeks, how often have you felt down, depressed or hopeless? Over the past two weeks, how often have you had little interest or pleasure in doing things? Over the past two weeks, how often have you felt tense, fidgety or restless? Over the past two weeks, how often have you felt tired or had little energy? The list of the 36 total features for this body system can be found in **Supplementary Table 2**.

Renal function: includes creatinine, urea and the electrolytes sodium and potassium, all derived from blood laboratory tests.

Medications: includes a total of 60 medication types, with self-reports relating to if the medication was taken between the study enrollment and the data point or not (1 or 0, respectively).

Medical conditions: 127 self reported medical conditions.

**Data Acquisition and Preliminary Filtering**

Participants completed a protocol of up to six movement assessment activities, captured using the Movement Lab (Newton VR Ltd., hereafter Newton Tech): treadmill walking at a fixed speed of 3 km/h (3 minutes), treadmill walking at a self-selected pace (1 minute), a 30-second sit-to-stand test, stationary walking (1 minute), and Romberg test with eyes closed for 30 seconds. Not all participants completed every activity; data gaps resulted from technical issues (data corruption) or safety protocols (e.g., exclusion of treadmill tasks for participants lacking appropriate footwear). The protocol was designed to probe complementary aspects of locomotor function, postural control, and functional capacity, with a total administration time of approximately 10 minutes per participant.

The Movement Lab was used in this instance with the Azure Kinect Body Tracking SDK (K4ABT), which provides 3D spatial coordinates (x, y, z) and confidence scores for 32 anatomical joints at a sampling rate of 30 frames per second (fps) (**see Extended Figure 3**). All recordings were performed from a frontal camera perspective. To ensure data fidelity we removed six distal joints with high variance and low detection confidence (HAND_LEFT, HANDTIP_LEFT, THUMB_LEFT, HAND_RIGHT, HANDTIP_RIGHT, and THUMB_RIGHT), retaining a 26-joint skeleton for analysis. These joints were identified empirically as having consistently low confidence scores, (i.e <10% of joints were above confidence 0) across the dataset.

We utilized a preprocessing pipeline designed to preserve the raw temporal dynamics of gait. Input sequences were fixed at 900 frames (~30 seconds at 30 fps) as most of the activities were 30 seconds long. To eliminate high-frequency shot noise, a temporal median filter with a window size of 3 frames was applied to each joint coordinate independently using a nearest-neighbor boundary mode. Spatial normalization was performed on a per-frame basis. Each skeleton was centered by subtracting the centroid computed from the pelvis joint (or pelvis and spine navel joints, depending on configuration). Following centering, the coordinates were scaled by the furthest joint distance from the origin within that frame, computed as the maximum Euclidean norm across all joints. This approach ensures that the spatial representation is invariant to the subject's absolute position and body size while strictly maintaining the original temporal structure and cadence of the walking sequence.

**Gait Feature Extraction**

Detailed gait analysis was performed on the 3D skeletal sequences by Newton Tech's Movement Lab - a commercial motion analysis pipeline. For walking tasks (treadmill and stationary walking), spatiotemporal parameters included cadence, bilateral step and stride length, step width, walking speed, swing and stance phase durations, double support time, and bilateral asymmetry indices for step length, step time, swing, and stance. Upper-body dynamics were captured through bilateral arm swing amplitude and head vertical sway. For postural tasks (stationary walking and Romberg test), center-of-mass sway was quantified using displacement metrics (anteroposterior and mediolateral distance, RMS distance, 95% confidence ellipse area), velocity metrics (mean and peak sway velocity), frequency-domain measures (mean frequency, total power, centroidal frequency, frequency dispersion), and derived indices including sway amplitude, normalized jerk, and sway path[26]. For the sit-to-stand task, extracted features included repetition count, chair-rise time, chair-rise time variability, peak reaction force normalized to body weight, and maximal rate of force development. In total, 71 unique parameter definitions computed across the five motor tasks yielded 672 activity-specific features. To reduce redundancy, we computed pairwise Pearson correlations across all features and applied hierarchical clustering with a correlation threshold of $|r| > 0.85$; a single representative feature was selected from each cluster, yielding 139 non-redundant descriptors. This target-agnostic reduction preserves the information content of the original feature set while eliminating collinear redundancy. These descriptors were combined with participant height to form the engineered baseline feature set used throughout this study (Full feature list and selected features are in **Supplementary Table 3**)

**Deep Learning Architecture and Pretraining**

The gait representation model is based on a Dual-Stream Spatio-Temporal Transformer (DSTformer) architecture (**Supplementary Figure 1**), inspired by the MotionBERT framework [27]. The model processes four-channel gait sequences (X, Y, Z, C), where the first three channels represent 3D joint coordinates and the fourth channel provides a per-joint detection confidence score from the pose estimation system. To capture long-range gait dynamics, the model was configured to process 900-frame trajectories (30 seconds of continuous movement at 30 fps). This was made computationally efficient through the integration of Rotary Position Embeddings (RoPE)[58] and the use of Flash Attention (via PyTorch's scaled_dot_product_attention) [59], in the temporal attention blocks enabling sequence length extrapolation without fixed positional embeddings, as well as being a better architecture for long sequence which achieved $O(n)$ memory complexity through chunked attention computation.The architecture consists of 8 transformer blocks with a feature dimension of 128 and 8 attention heads. Each block features parallel spatial-temporal (ST) and temporal-spatial (TS) attention streams that are integrated via learned fusion weights, allowing the model to adaptively balance the contribution of spatial joint relationships and temporal movement dynamics. The spatial attention operates across joints within each frame with absolute Positional Encoding, while the temporal attention operates across frames for each joint, with RoPE applied only to the temporal dimension to encode relative position information. For sequence reconstruction, we employed a decoder consisting of 3 Linear Layer and Tanh activation that maps the latent embeddings back to the original 3D coordinate space.

**Data Augmentation**

To enhance the model's generalization capabilities and account for sensor noise and natural movement variability, we implemented a data augmentation strategy using Gaussian jittering. Additive white Gaussian noise was applied to the joint coordinates, sampled from a normal distribution with a standard deviation of σ=0.05 (in normalized coordinate space). This augmentation effectively simulates the inherent precision limitations of the pose-estimation system without distorting the underlying gait patterns. A hybrid training protocol was employed wherein the model was exposed to both the original (clean) and the augmented (noisy) sequences. This dual-exposure strategy allowed the network to simultaneously learn noise-robust features and high-fidelity motion reconstruction.

**Pretraining**

Pretraining was conducted using a Denoising Masked Autoencoder (DMAE) objective. In this setup, joint positions were stochastically masked by zeroing out their input channels according to the multi-level masking strategy described above, forcing the model to reconstruct the missing spatio-temporal information. When adding noise augmentation we added 0.05 units of gaussian noise. The optimization objective combined Euclidean distance reconstruction loss (MPJPE) with a 3D velocity loss with $\lambda = 0.1$ that penalizes discrepancies in frame-to-frame joint velocities, and scaled Mean Per-Joint Position Error (NMPJPE) with $\lambda = 0.1$ to ensure anatomically consistent reconstructions.

$$Loss = MPJPE + \lambda_1 NMPJPE + \lambda_2 VelocityError$$

All models were trained on 1 NVIDIA L40S GPU using PyTorch[60] with the AdamW optimizer[61] (weight decay 0.01), a cosine annealing learning rate schedule[62], and linear warmup over the first 0.3 epochs. To maximize hardware utilization, we utilized mixed-precision training[63] and torch.compile graph optimization for accelerated computation. We reserved 10% of the data for evaluation; the model achieved nearly identical reconstruction performance on this held-out set compared to the training set, demonstrating strong generalization and the successful preservation of high-fidelity gait characteristics within the learned embedding space.

**Masking Strategy**

The model was pretrained using a Masked Autoencoder (MAE) framework with a composite masking strategy designed to achieve a high effective masking ratio and encourage the learning of robust structural and temporal embeddings. The masking strategy combined:

1. Anatomical Group Masking: Joints were partitioned into six anatomical regions: Left Leg (hip, knee, ankle, foot), Right Leg, Left Arm (clavicle, shoulder, elbow, wrist), Right Arm, Torso

(pelvis, spine navel, spine chest, neck), and Head (head, nose, eyes, ears). During training, 4 out of 6 regions were randomly selected and masked across temporal spans of 16 consecutive frames. This spatial-temporal span masking forced the model to reconstruct limb-specific trajectories based on the global coordination of the remaining body segments.

2. Random Frame Masking: Entire frames were randomly selected and masked with a probability of 5% across the sequence, forcing the model to interpolate missing temporal information from preceding and succeeding movement states.

The combination of these masking strategies results in an effective masking ratio of approximately 70% of the input signal, creating a challenging reconstruction objective that promotes the learning of rich, generalizable representations.

**Hierarchical Pooling and Feature Aggregation**

To transform variable-length kinematic sequences into standardized physiological representations, we benchmarked five hierarchical pooling architectures on age prediction with 5-fold cross validation. The objective was to identify an aggregation strategy that balances the preservation of anatomical structure with robustness to temporal variability. We evaluated strategies ranging from baseline global statistics to fine-grained joint-specific embeddings. As detailed in **Extended Table 2**, the analysis indicated that increasing spatial granularity does not strictly correlate with predictive performance. For instance, maintaining independent representations for all 26 joints (Variant 4) or strictly separating bilateral limbs (Variant 3) resulted in lower correlations, likely due to overfitting local noise or sensitivity to minor gait asymmetries. The selected architecture, Variant 5 (Bilaterally-Merged Groups with Percentile Pooling), demonstrated the highest Pearson correlation (r=0.5765). This strategy aggregates skeletal data into four functional domains- Head, Torso, Arms, and Legs- by merging bilateral limb data to reduce asymmetry-induced artifacts. Furthermore, by utilizing the 99th percentile alongside the temporal mean, Variant 5 captures both steady-state dynamics and transient peak activations (e.g., heel-strike impact), yielding a robust 1024-dimensional physiological signature.

**Downstream Evaluation Framework and Statistical Analysis**

**Nested Cross-Validation Strategy.** To ensure rigorous evaluation and account for the longitudinal structure of the dataset, we implemented a nested, subject-level cross-validation (CV) framework. Data splitting was performed strictly at the subject level to prevent data leakage between visits. We employed a 5-fold outer loop for performance estimation, repeated across 15 distinct random seeds (generating n=75 independent train-test splits) to produce stable mean estimates and confidence intervals. For binary targets with significant class imbalance (<20% minority class), stratified splitting was applied to maintain consistent class distributions.

**Model Selection and Hyperparameter Optimization.** Within the inner CV loop (4-fold), we decoupled model selection from performance evaluation. We competitively evaluated linear models (Ridge regression for continuous targets; Logistic regression for binary/ordinal targets) against nonlinear gradient-boosted decision trees (LightGBM)[64]. Hyperparameters (e.g., regularization strength, tree depth, learning rate) were optimized via RandomizedSearchCV. The optimal model architecture was selected automatically based on inner-loop performance, ensuring that the outer-loop test data remained unseen during the optimization process.

**Ensemble Architecture.** To integrate evidence from diverse movement contexts, we utilized a unified ensemble approach. The feature matrix was constructed by concatenating the 1,024-dimensional latent embeddings with activity-specific metadata (activity type and sequence index). Categorical metadata were one-hot encoded for linear models and ordinally encoded for tree-based models. Final predictions were generated using a late-fusion strategy: model outputs were averaged across all gait activities per subject-visit pair to produce a single consensus prediction.

**Direct Prediction Analysis.** To evaluate the raw predictive capacity of gait embeddings without demographic or anthropometric adjustment, we trained models using gait embeddings as the sole input features. For each of the 3,210 phenotypic targets, predictions were evaluated using 5-fold nested cross-validation repeated across 15 random seeds. Statistical significance was assessed using the Pearson correlation p-value for each target, with correction for multiple comparisons using the Benjamini–Hochberg FDR procedure at $q < 0.05$. This analysis simulates a passive screening scenario in which only a gait recording is available, with no accompanying clinical information (Figures 3, Extended Figure 1).

**Covariates and Baseline Definitions.** To isolate the specific predictive utility of gait physiology beyond standard demographic and anthropometric factors, we utilized Age, Gender, Body Mass Index (BMI), and Visceral Adipose Tissue (VAT) as covariates. For all predictive tasks, we constructed a null baseline model trained solely on these covariates. This design allows for the explicit quantification of the added information value provided by the gait embeddings.

**Inference and Statistical Validation.** The final predictive power for each model was reported as the median metric (Pearson r for regression; AUC-ROC for classification) across the seeds of predicted folds. To rigorously validate associations, we compared the predictive performance of the full model (Gait + Covariates) against the baseline null model. For each target, we performed a two-sided t-test comparing the distribution of scores from the full model (n=15) against the distribution of scores from the baseline model (n=15). A gait association was considered statistically significant only if the full model's performance distribution was significantly higher than that of the baseline after BH-FDR correction (FDR $q<0.1$).

**Evaluation Metrics.** Performance was assessed using metrics appropriate for the statistical properties of each biomarker: Pearson correlation coefficient (r) and coefficient of determination ($R^2$) for continuous regression targets; Area Under the Receiver Operating Characteristic Curve (AUC-ROC) for binary classification; and Spearman rank correlation ($\rho$) for ordinal variables. Given the high discriminative power of gait features for sex (AUC ≈ 0.99), analyses were performed on both the full cohort and stratified by sex to detect sex-specific physiological associations.

**Association with current diseases and medical conditions** To identify associations between gait embeddings and diseases, medical conditions or medication intake, we applied a logistic regression model with a ridge regularization (using the scikit-learn [65] Python package), for male and female participants separately. Inference and validation were performed as mentioned above, with the exception of the model score calculation, which was computed as the AUC for each iteration, to adapt to binary classification.

**Multiple Testing Correction.** To address the multiple comparisons problem inherent in high-dimensional biomarker analysis, statistical significance was determined using the Benjamini-Hochberg False Discovery Rate (FDR) procedure. Correction was applied independently for each feature group (hypothesis family). [66]

### Anatomical Interpretability and Feature Attribution

To decipher which biomechanical components drive specific physiological predictions, we developed a post-hoc interpretability framework based on systematic joint perturbation. This analysis was conducted on a frozen backbone.

Ablation analyses were performed exclusively on embeddings derived from treadmill walking activities (3 km/h and self-selected pace). Activities with fundamentally different biomechanical demands (sit-to-stand, Romberg test, stationary walking) were excluded from this analysis, as the relative contribution of each joint group to task execution differs inherently across motor paradigms, confounding the attribution of importance to health-related signals. [67]

**Anatomical Partitioning** We partitioned the 26-joint skeleton into four functional domains based on anatomical adjacency:

- **Head:** Joints 20–25 (Head, nose, eyes, ears).
- **Torso:** Joints 0–3 (Pelvis, spine navel, spine chest, neck).
- **Arms:** Joints 4–11 (Bilateral clavicles, shoulders, elbows, wrists).
- **Legs:** Joints 12–19 (Bilateral hips, knees, ankles, feet).

**Perturbation Strategy** For each target biomarker and each anatomical group g, we computed performance metrics under two distinct masking conditions:

**1. Masked Drop ($\Delta_{drop}$):** The joints in group g were zeroed out while retaining the rest of the body. This measures the *necessity* of the region (i.e., how much performance degrades in its absence).

$$\Delta_{drop} = S_{baseline} - S_{(Masked)}$$

**2. Isolation Score ($r_{solo}$):** All joints *except* those in group g were zeroed out ("All-But" strategy). This measures the *sufficiency* of the region (i.e., how much predictive signal the region contains independently).

$$r_{solo} = S_{(Isolated)}$$

**Importance Calculation** A direct summation of these metrics is problematic because $\Delta_{drop}$ represents small relative deltas (typically -0.05 to 0.17), whereas $r_{solo}$ represents absolute performance scores

(typically 0.1 to 0.7). To prevent the magnitude of the isolation score from dominating the attribution, we implemented a dual-normalization strategy.

Let $D$ be the set of drop scores for all four groups and $R$ be the set of isolation scores. We first Min-Max normalized each component independently across the four groups:

$$\hat{\Delta}_{drop} = (\Delta_{drop} - min(D)) / (max(D) - min(D)) \quad \hat{r}_{solo} = (R_{solo} - min(R)) / (max(R) - min(R))$$

The final importance score $I_g$ was computed as the unweighted sum of these normalized components, followed by a final scaling to the unit interval [0, 1] for visualization:

$$I_g = Norm(\hat{\Delta}_{drop} + \hat{r}_{solo})$$

This composite metric ensures that a body region is deemed "important" only if it is both necessary for peak performance and sufficient to drive prediction independently.

**Methods References**

**Data Availability**

Data in this paper are part of the Human Phenotype Project (HPP) and are accessible to researchers from universities and other research institutions at https://humanphenotypeproject.org/data-access.

The HPP data include personal information and, in compliance with institutional review board regulations, cannot be made publicly available. Interested bona fide researchers should contact info@pheno.ai to obtain instructions for accessing the data, which is typically granted within a few days.

**Code Availability**

Code used in this study will be available at https://github.com/AdamGabet/GaitPredict (access available upon request during review).

**Acknowledgements**

We thank A. Gefen, A. Shapiro, I. Efanov and V. Efanov from Newton VR LTD for help on this project, E. Barkan for advice and H. Saranga for meaningful conversation and all the Segal Lab for the support and feedback.

**Contributions**

A.G. conceived the project, designed the DL model and conducted all analyses, interpreted the results and wrote the manuscript. S.K., G.L., G.S and A.Z. reviewed the results and the manuscript. S.G. interpreted the results and reviewed the manuscript. A.W. developed protocols and oversaw sample collection and processing. Y.B., A.R. and O.D. collected the data and reviewed the manuscript. N.G. and D.K. Performed data preprocessing. R.S. collected and annotated data. E.S. conceived and directed the project and analyses, designed the analyses and reviewed the results and the manuscript.

**Competing Interests**

A.R., O.D., and Y.B. are employees of Newton VR Ltd., which developed the Movement Lab system used for gait data acquisition in this study. A.R. is also affiliated with BioPilot AI. A.W. and E.S. are paid consultants of Pheno.AI, Ltd. All other authors declare no competing interests.

# Extended Data

## Extended Table 1

| Characteristic, mean (std) | Demographics | | Treadmill walking 3KPH (5400s) | | Stationary Walk (1800s) | | Treadmill Walking at Self Selected Pace (1800s) | | Romberg's test eyes closed (900s) | | Sit to Stand (900s) | |
|---|---|---|---|---|---|---|---|---|---|---|---|---|
| | Age | BMI | Cadence (steps/min) | Stride Length (m) | 95% Ellipse Area (mm²) | Sway Amplitude AP (mm) | Cadence (steps/min) | Stride Length (m) | 95% Ellipse Area (mm²) | Sway Amplitude AP (mm) | Repetitions (count) | Chair-Rise Time (ms) |
| All (3,414) | 52.48 (10.10) | 25.74 (3.79) | 97.60 (10.00) | 1.03 (0.10) | 63.22 (38.70) | 26.17 (8.92) | 100.36 (12.24) | 1.05 (0.18) | 0.35 (0.34) | 294.42 (106.82) | 9.87 (2.33) | 689.5 (108.9) |
| Male (1,652) | 52.07 (9.98) | 26.18 (3.28) | 94.41 (8.40) | 1.06 (0.09) | 62.24 (37.36) | 24.93 (8.41) | 99.02 (11.74) | 1.11 (0.18) | 0.38 (0.39) | 307.06 (98.82) | 9.72 (2.36) | 715.6 (111.7) |
| Female (1,762) | 52.87 (10.21) | 25.34 (4.17) | 100.76 (10.45) | 0.99 (0.10) | 64.17 (39.94) | 27.36 (9.23) | 101.67 (12.58) | 0.99 (0.17) | 0.33 (0.29) | 282.22 (99.83) | 10.01 (2.29) | 664.1 (99.9) |

## Extended Table 2

| Pooling Strategy | Feature Dim. | Age prediction (Pearson r) | Description & Rationale |
|---|---|---|---|
| V1: Mean-Max Global | 256 | 0.618 | **Baseline.** Flattens spatio-temporal features using global mean and max. Computationally efficient but loses structural hierarchy. |
| V2: Hierarchical Stats | 384 | 0.64 | Two-stage. 1. Pooling joints with mean, max, std then mean pooling across time. |
| V3: Anatomical Groups | 768 | 0.659 | Pool with mean across time and joints within 6 regions, maintaining separate left/right limb representations. |
| **V4: Per-Joint Pre-logits** | 832 | 0.605 | Independent pooling for all 26 joints using embeddings after final MLP before reconstruction layer. Lowest performance suggests overfitting to local noise. |
| **V5: Merged Groups + %** | **1024** | 0.691 | **Selected Method.** Pool with mean and 99th percentile across time and then 4 joint groups. Merges bilateral limbs to reduce noise. |

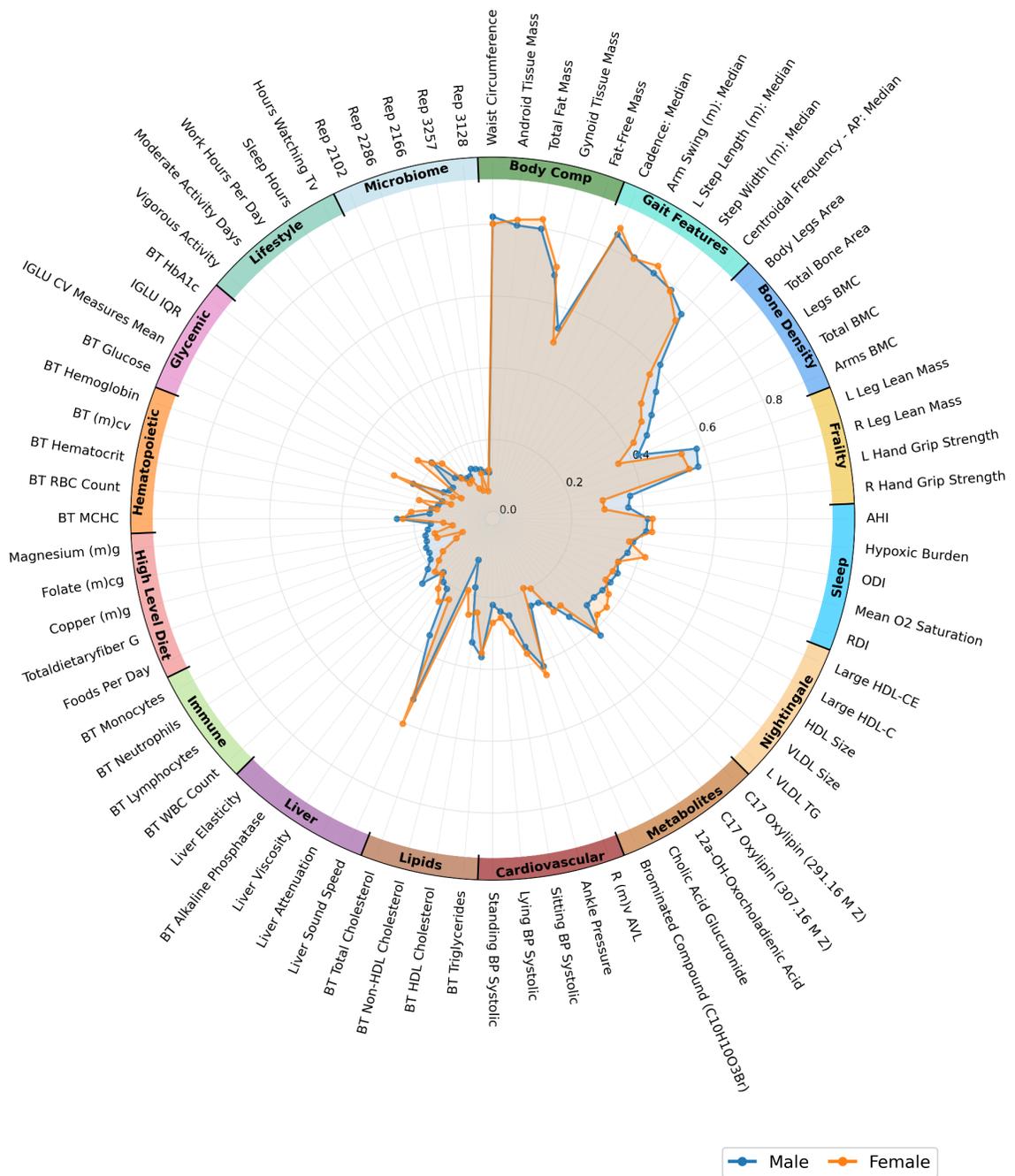

**Extended Figure 1: Sex-stratified gait embedding predictions across body systems.** Radar plot showing the top 5 features most strongly predicted by the Gait Fusion model for each body system, stratified by sex (male, blue; female, orange). Each slice represents a single phenotype, grouped by body system (labeled in the inner ring). The radial axis spans Pearson *r* correlation of predictions from 0 to 1. Features within each body system are ordered clockwise. Predictive performance is largely concordant between sexes for structural and metabolic targets (body composition, bone density, lipids), while greater divergence is evident for hematopoietic, lifestyle, and sleep phenotypes. Predictions were generated using linear ridge regression on gait embeddings from the frozen GaitMAE encoder, evaluated via 5-fold nested cross-validation repeated across 15 random seeds ($n = 75$ train–test splits). Of the 3,210 sex-stratified phenotype predictions tested, 1,098 were significant in males and 1,205 in females after FDR correction. No demographic or anthropometric covariates were included, simulating a passive screening scenario in which only a gait recording is available. AHI, apnea–hypopnea index; BMC, bone mineral content; BT, blood test; HDL-C, high-density lipoprotein cholesterol; HDL-CE, high-density lipoprotein cholesterol ester; IGLU, indices of glycemic variability; MCHC, mean corpuscular hemoglobin concentration; ODI, oxygen desaturation index; RBC, red blood cells; RDI, respiratory disturbance index; VLDL, very-low-density lipoprotein; WBC, white blood cells.

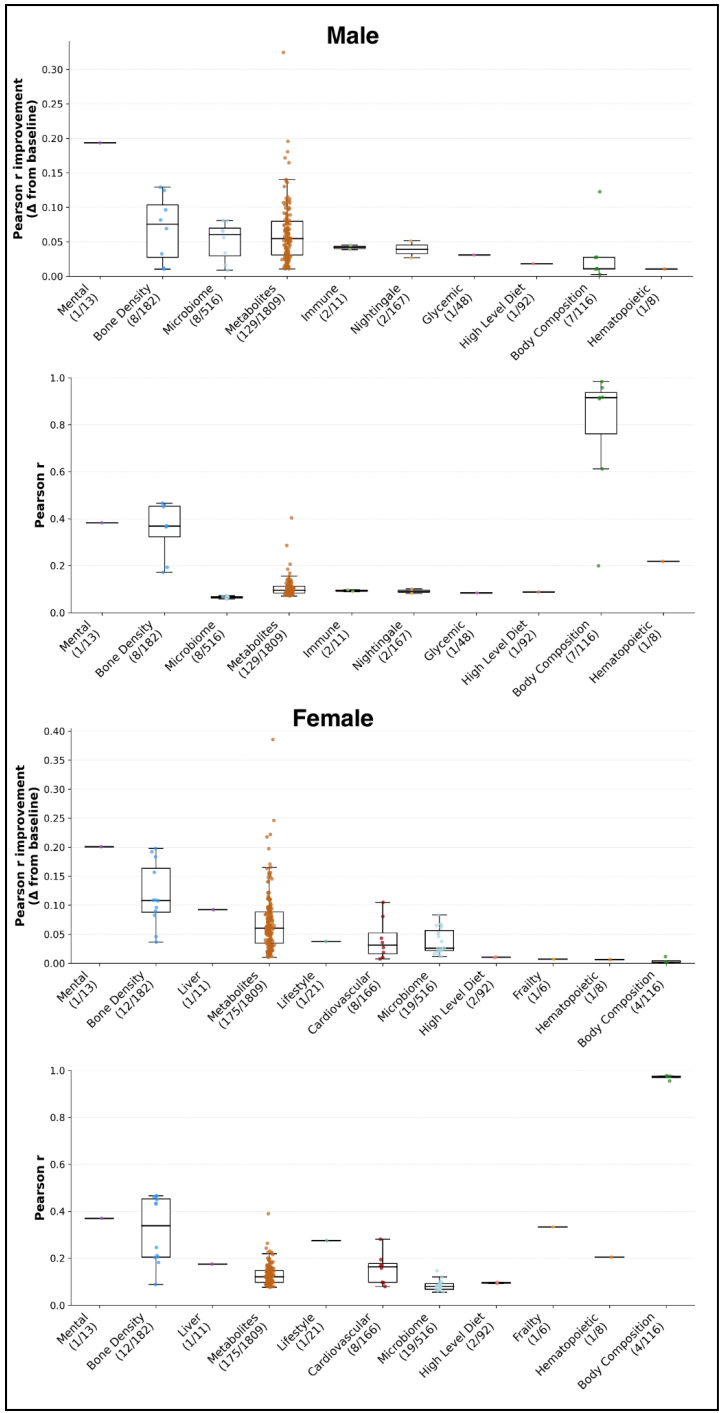

**Extended Figure 2: Traditional gait descriptors provide limited predictive gain beyond demographic and anthropometric baselines.** Box and swarm plots showing the predictive performance of engineered gait features (cadence, stride length, sway metrics, and height) beyond a demographic baseline (age, BMI, VAT, and height) for male (top) and female (bottom) participants. Top panels show the predictive gain (Δr) attributable to engineered gait features; bottom panels show the median Pearson r

of the combined model. Only body systems containing at least one significantly improved feature are shown. Of 3,210 sex-stratified phenotype predictions tested, 160 were significant in males and 225 in females (FDR-corrected P < 0.1), compared with 697 and 611 respectively for learned gait embeddings (Figure 4). Predictions were evaluated using 5-fold nested cross-validation repeated across 15 random seeds (n = 75 train–test splits). Significance was assessed by two-sided t-test comparing full versus baseline model performance distributions.

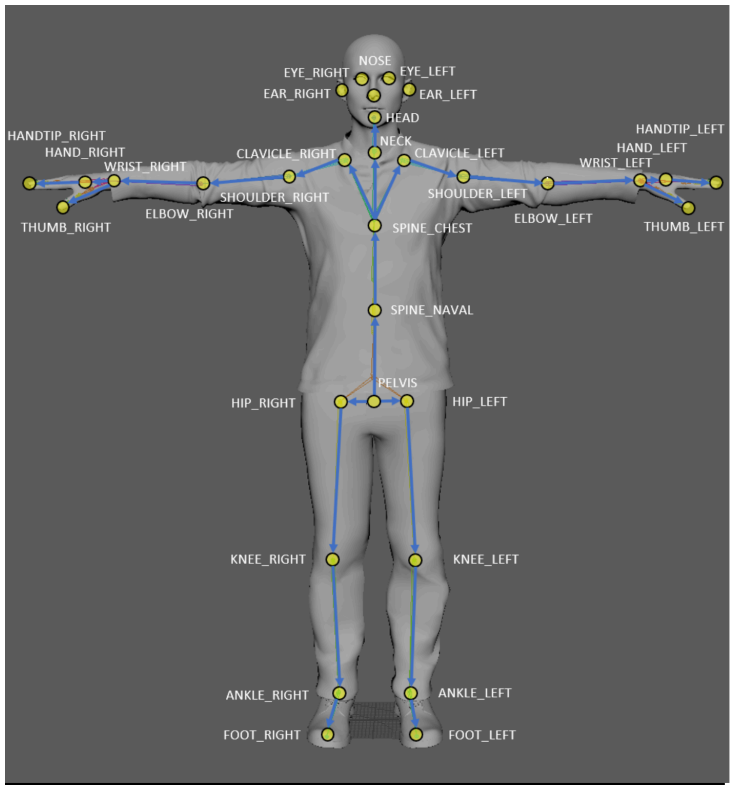

**Extended Figure 3: Azure Kinect Body Tracking skeleton model.** The 32-joint skeleton output by the Azure Kinect Body Tracking SDK (K4ABT), shown on a reference mannequin. Yellow nodes denote tracked joints; blue arrows indicate the kinematic chain. Six distal hand joints (HAND, HANDTIP, THUMB, bilateral) were excluded due to low detection confidence, yielding a 26-joint skeleton for analysis.

**Supplementary**

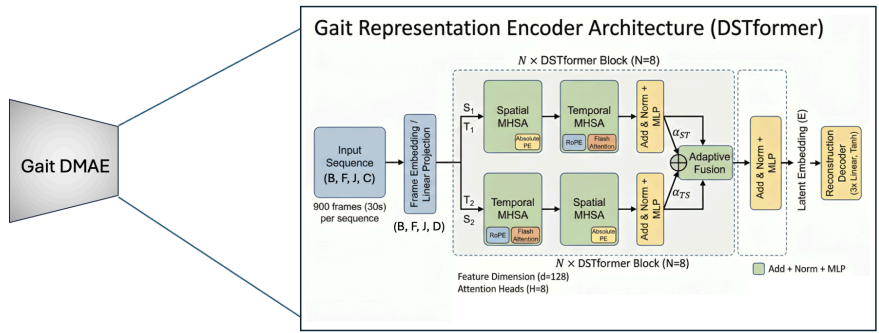

**Supplementary Figure 1 | Detailed architecture of the Gait Representation Encoder (DSTformer).** The Dual-Stream Spatiotemporal Transformer (DSTformer, 3.2M parameters) processes input skeleton sequences of shape (B, F, J, C), batch size, 900 frames (30 s at 30 fps), 26 joints, and 4 channels (x, y, z coordinates plus detection confidence), through a frame embedding and linear projection layer, yielding a (B, F, J, D) tensor with feature dimension D = 128. The architecture comprises N = 8 repeated DSTformer blocks, each containing two parallel attention streams: a spatial-first stream that applies spatial multi-head self-attention (MHSA) with absolute positional encoding across joints before temporal MHSA with Rotary Position Embeddings (RoPE) and Flash Attention across frames, and a temporal-first stream that reverses this order. Stream outputs are combined via learned adaptive fusion weights ($\alpha\_ST$, $\alpha\_TS$), followed by residual addition, layer normalization, and a feed-forward MLP. The fused representation passes through a final normalization and MLP layer to produce latent embeddings (E), which serve as input to downstream phenotype prediction models. A reconstruction decoder consisting of three linear layers with Tanh activation maps embeddings back to the original coordinate space during self-supervised pretraining. All attention heads H = 8.

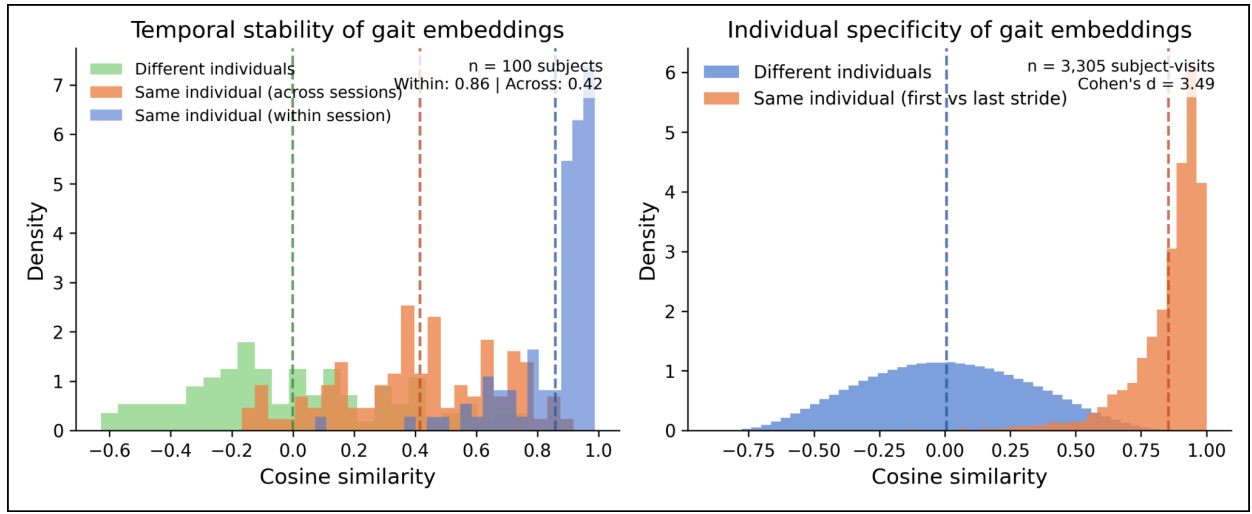

**Supplementary Figure 2 | Gait embeddings encode individual identity and are temporally stable. a,** Temporal stability of gait embeddings in repeat visitors (n = 100 subjects). Within-session similarity (first versus last stride, blue) remains high (mean = 0.86), while across-session similarity (last stride of session 1 versus first stride of session 2, orange) is lower (mean = 0.42) yet well above the random inter-individual baseline (green, centered near zero). Dashed vertical lines indicate distribution means. **b,** Individual specificity of gait embeddings across the full cohort (n = 3,305 subject-visits). Distributions of cosine similarity between the first and last stride embeddings of the same individual within a session (orange) versus embeddings from different individuals (blue). Cohen's d = 3.49 indicates strong separation, confirming that learned representations capture individual-specific gait signatures. These results demonstrate that individual gait signatures persist across visits despite the expected increase in variability over longer time intervals.

**Supplementary table 1-3**

📊 **supplementary.xlsx**